\documentclass[twocolumn, journal]{IEEEtran}

\usepackage[switch]{lineno}
\usepackage{amsmath}
\usepackage[pdftex]{graphicx}
\usepackage[font=small,labelfont=bf]{caption}
\usepackage[colorlinks,allcolors=blue]{hyperref}
\ifCLASSINFOpdf
\else
\fi

\begin{document}
\title{CrackGAN: Pavement Crack Detection Using Partially Accurate Ground Truths Based on Generative Adversarial Learning}

\author{\IEEEauthorblockN{Kaige~Zhang, Yingtao Zhang, and H. D. Cheng}
\thanks{*Accepted by IEEE Transactions on Intelligent Transportation Systems.   }% <-this % stops a space
%\thanks{Yingtao~Zhang is with Dept. of Computer Science, Harbin Institute of Tech., Harbin, China 150001. E-mail: yingtao@hit.edu.cn.}% <-this % stops a space
%\thanks{H. D.~Cheng, corresponding author, is with Dept. of Computer Science, Utah State University, Logan, USA 84322. Email: hd.cheng@aggiemail.usu.edu.}% <-this % stops a space
}


\markboth{IEEE T-ITS \LaTeX\ Class Files,~Vol.~XX, No.~XX, Dec.~2018}%
{}



\maketitle
%\linenumbers

\begin{abstract}
Fully convolutional network is a powerful tool for per-pixel semantic segmentation/detection. However, it is problematic when coping with crack detection using partially accurate ground truths (GTs): the network may easily converge to the status that treats all the pixels as background (BG) and still achieves a very good loss, named ``All Black'' phenomenon, due to the unavailability of accurate GTs and the data imbalance. To tackle this problem, we propose crack-patch-only (CPO) supervised generative adversarial learning for end-to-end training, which forces the network to always produce crack-GT images while reserves both crack and BG-image translation abilities by feeding a larger-size crack image into an asymmetric U-shape generator to overcome the ``All Black'' issue. The proposed approach is validated using four crack datasets; and achieves state-of-the-art performance comparing with that of the recently published works in efficiency and accuracy.
\end{abstract}

\begin{IEEEkeywords}
Pavement crack detection, fully convolutional networks, generative adversarial learning, and partially accurate GTs.
\end{IEEEkeywords}

% \fi
\IEEEpeerreviewmaketitle

\section{Introduction}
\IEEEPARstart{A}{utomatic} pavement crack detection is a challenging task in intelligent pavement surface inspection system \cite{rf1}. It is also a research topic for more than three decades. However, industry-level pavement crack detection task is still not well solved: many published references have reported good results on specific crack datasets \cite{rf3}; however, the methods failed when processing industrial pavement images of which the cracks were thin and the precise pixel-level ground truths (GTs) were difficult to obtain \cite{rf4, rf5}. Recently, fully convolutional network (FCN) \cite{rf6}, trained in end-to-end for pixel-level object segmentation/detection, was applied to pavement crack detection \cite{rf5, rf7}. However, it suffered from the ``All Black'' issue when processing industrial images: the network converged to the status that treated all the pixels as background (BG) \cite{rf5}; and similar issue was also reported in \cite{rf7} where the FCN failed to detect thin cracks. 

It is known that deep learning is a data driven approach which heavily relies on the training data with accurate GTs. Due to the domain sensitivity (i.e., the performance of a ``well-trained'' network may decrease when utilizing the datasets obtained from different road sections and/or during different periods), it is necessary to \textit{manually} mark the GTs to re-train the models for new pavement crack detection tasks.  In industry, the pavement images are captured using a camera mounted on top of a vehicle running on the road. Under such setting, most cracks are very thin and crack boundaries are vague, which makes the annotation of pixel-level GTs very difficult. Instead of the labor-intensive per-pixel crack annotation, marking the cracks as 1-pixel curves is more feasible and preferable in practice because of its simplicity and low labor-cost, and such GT is named \textit{labor-light} GT. However, such GTs may not completely match the cracks at pixel-level accurately; i.e., they are partially accurate GTs, and that makes the loss computation inaccurate. Moreover, as a long-narrow target, a crack can only occupies a very small area in a full image. Since patch-wise training is equivalent to loss sampling in FCN \cite{rf6}, directly training FCN for pixel-level crack detection makes the training set heavily imbalanced; moreover, such problem cannot be handled by simply rebalancing the data via loss function since the GTs are not accurate. The observation is that the network will simply converge to the status that treats the entire crack image as BG (labeled with zero), and still can achieve a good detection accuracy (BG-samples dominate the accuracy calculation). It is named ``All Black'' problem which is pretty common in industrial pixel-level pavement crack detection.
 
In general, the existing computer vision-based crack detection approaches could be grouped into two categories: rule-based and machine learning-based methods. Rule-based methods try to extract some pre-defined features to identify the cracks. Cheng et al. \cite{rf8} proposed a fuzzy logic-based intensity thresholding method for crack segmentation based on the assumption that crack pixels were darker than BG pixels; however, the method failed when processing crack images with low foreground-background contrast. Wang et al. \cite{rf9} introduced a wavelet-based edge detection algorithm, and the drawback was that it could not handle the cracks with high curvature or low continuity well. Oliveira et al. \cite{rf10} proposed a dynamic thresholding method for crack detection based on information entropy, which was sensitive to noise. Zou et al. \cite{rf11} designed an intensity-difference measuring function to find an optimal threshold for crack segmentation; however, the robustness was poor and the method was easy to fail when working on different datasets. Many works introduced some crack linking method to enhance the crack continuity \cite{rf12}-\cite{rf17}. However, these methods did not solve the problem well and usually produced intolerable false positives for linking together the noises. In addition, Tsai et al. \cite{rf18} performed a comprehensive study on the performances of six low-level image segmentation algorithms, and Abdel-Qader et al. \cite{rf19} discussed different edge detectors, including Sobel, Canny, and fast Haar transformation \cite{rf20}. The rule-based approaches are easy to implement; however, they are sensitive to noise, which results in poor generalizability.

Machine learning-based methods have attracted increasing attentions during the past two decades. These methods perform crack detection following two steps: feature extraction and pattern classification. Cheng et al. \cite{rf21} and Oliveira et al. \cite{rf22} utilized mean and variance of an image block as the features to train classifiers for pavement crack detection. However, the good performances heavily relied on complex post processing.  Hu et al. \cite{rf23} and Gavilan et al. \cite{rf24} utilized textural information to set up the feature vectors and employed support vector machine (SVM) for the classification. However, they could not handle the problem well when processing the images with complex pavement textures. Zalama et al. \cite{rf25} employed Gabor filters for feature extraction and AdaBoosting \cite{rf25} for crack identification; and Shi et al. \cite{rf3} combined multi-channel information to set up the feature vector, and employed random structure forest \cite{rf26} for crack-token mapping. These methods tried to solve the problem by extracting some hand-crafted features and training a classifier to discriminate cracks from the noisy BG; however, they did not address the issue well because the hand-crafted feature descriptors usually calculated statistics locally and lacked good global view, even the statistics from different locations were combined together. Thus, they could not represent the global structural pattern well which was important to discriminate cracks from the noisy textures.

As one of the most important branches in machine learning, deep learning has achieved great success during the past ten years, and it is the most promising way to solve challenging object detection problems, including pavement crack detection. Initially, deep learning-based object detection methods relied on window-sliding or region-proposal; and these methods tried to find a bounding box for each possible object in an image. R-CNN (region-based convolutional neural networks) \cite{rf27} was the early work which utilized selective search \cite{rf28} to generate candidate regions, and then sent the regions into a CNN for classification. Based on R-CNN, Cha et al. \cite{rf29} designed a convolutional network for pavement crack detection which worked with window-sliding mode. Zhang et al. \cite{rf4} employed a CNN for pre-classification which removed most of the noise areas before performing crack and sealed crack detection. Problems of these methods were: (1) window-sliding-based strategy was impractical due to the huge time complexity, especially when processing large images \cite{rf5}; (2) traditional region-proposal methods \cite{rf28} were unable to select good candidate regions from the noisy pavement images, and it was also inefficient because a great number of candidate regions had to be processed for a full-size image. Zhang et al. \cite{rf30} employed parallel processing to improve the computation efficiency of region-based methods; however, the computation and resource costs were expensive. Zhang et al. \cite{rf5} addressed the computational issue by generalizing a classification network to an end-to-end detection network which minimized the redundant convolutional operations. FCN is a one-stage pixel-level semantic segmentation method without window-sliding. Recently, Yang et al. \cite{rf7} employed FCN for pixel-level crack detection and achieved good results on concrete-wall images and pavement images with clear cracks; the method failed to detect thin cracks. Moreover, the method relied on accurate pixel-level GTs which were labor-intensive and often infeasible under industrial setting. In addition, deep learning-based crack detection articles have been keeping on appearing. Chen et al. \cite{rf31} and Park et al. \cite{rf32} proposed NB-CNN and patch-CNN for crack detection, respectively; however, the networks could only process fixed input size images which limited the practical application. Tong et al. \cite{rf33} utilized DCNN for crack length estimation; and Hoang et al. \cite{rf34} employed CNN and edge detector for crack recognition. Gopalakrishnan et al. \cite{rf35} used transfer learning for pavement distress detection with a DCNN. Zou et al. \cite{rf36} and Yang et al. \cite{rf37} introduced DCNNs for crack detection with hierarchical feature learning. The methods were either based on the traditional classification network with fully connected layers which only could handle fixed input-size images, or based on the FCN architecture which relied on the accurate, labor-intensive GTs.

In this paper, we propose CrackGAN for pavement crack detection with the following contributions: (1) it solves a practical and essential problem,``All Black'' issue, existing in deep learning-based pixel-level crack detection methods; (2) it proposes the crack-patch-only (CPO) supervised adversarial learning and the asymmetric U-Net architecture to perform the end-to-end training; (3) the network can be trained with partially accurate GTs generated by labor-light method which can reduce the workload of preparing GTs significantly; (4) furthermore, it can solve data imbalance problem which is the byproduct of the proposed approach. Moreover, even the network is trained with small image patches and partially accurate GTs, it can deal with full-size images and achieve great performance.

The rest of the paper is organized as follows: In section II, it discusses the related works. In section III, it introduces the proposed method. In section IV, it describes the evaluation metrics and the experimental results. At the end, it provides the conclusion.

\section{Related Works}
In this section, it discusses the techniques related to the proposed method.
\subsection{Generative Adversarial Networks}
Goodfellow et al. \cite{rf38} proposed generative adversarial network (GAN) which could be trained to generate real-like images by conducting a max-min two-player game. Based on GAN, Mirza et al. \cite{rf39} proposed conditional GAN which introduced additional information (the condition) to the generator for producing specific outputs according to the input condition. While GAN is difficult to train, Radford et al. \cite{rf41} proposed deep convolutional generative adversarial network (DC-GAN) which configured the generator with convolutional layers, and the training became easier and more stable. Based on conditional GAN, Isola et al. \cite{rf40} set up the generator with an encoding-decoding network, then the GAN became an image-to-image translation network. Inspired by these works, we formulate the crack detection as an image-to-image translation problem, and introduce generative adversarial loss to regularize the objective function to overcome the ``All Black'' issue using partially accurate GTs generated by labor-light method.

%\IEEEpubidadjcol
\subsection{Transfer Learning in DCNN}
Transfer learning has been widely used for training deep convolutional neural networks, which intends to transfer knowledge learned in previous tasks to make the training easier \cite{rf43}. Depending on situations, there are different transfer learning strategies according to ``what knowledge to transfer'' and ``how to transfer the knowledge''. Yosinski et al. \cite{rf44} discussed the knowledge transferability of different layers in deep neural networks. Oquab et al. \cite{rf45} transferred the mid-level knowledge for nature image processing. Zhang et al. \cite{rf4} transferred the generic knowledge learned from ImageNet \cite{rf46} to ease the training of a crack detection network. Zhang et al. \cite{rf5} also transferred the mid-level knowledge via introducing a dense-dilation layer into FCN to improve crack localization accuracy. The proposed approach employed transfer learning to train the prototype of the encoding network, and also transferred the knowledge from a pre-trained DC-GAN to provide the generative adversarial loss for the end-to-end training.

\subsection{Fully Convolutional Network}
Regular DCNN usually employed convolutional layers for feature extraction and fully connected layers for classification \cite{rf47}. Interestingly, it turned out that the fully connected layer could be considered as a special case of the convolutional layer with kernel size equal to the input size \cite{rf5}.  Long et al. \cite{rf6} proposed the fully convolutional network (FCN) for per-pixel semantic segmentation. Based on FCN, Chen et al. proposed DeepLab model \cite{rf48} for multi-scale semantic segmentation; Ronneberger et al. \cite{rf49} proposed U-Net architecture for medical image segmentation. Xie et al. \cite{rf50} employed FCN for contour detection; Yu et al. \cite{rf51} proposed dilated convolutional design for multi-scale context aggregation. To improve the computation efficiency, Zhang et al. \cite{rf5} generalized a patch-based classification network to be a detection network for crack detection where FCN was employed. The proposed approach introduces FCN to extend the U-Net to the asymmetric U-Net, which provided the network with the translation ability of both crack and BG images. The FCN design also enables the patch-based CrackGAN (trained with small image patches) to work on the full-size images seamlessly.

\section{Proposed Method}
Fig.1 is the overview of the proposed method. \textit{D} is a pre-trained discriminator obtained directly from a pre-trained DC-GAN using crack-GT patches only. Such pre-trained discriminator will force the network to always generate crack-GT images, which is the most important factor to overcome the ``All Black'' issue. The pixel-level loss is employed to ensure that the generated crack patterns are the same as that of the input patch via optimizing the L1 distance based on the dilated GTs. Since the network is trained with crack-patch only, the asymmetric U-shape architecture is introduced to enable the translation abilities of both crack and non-crack images. After training, the generator itself will serve as the crack detection network. In addition, the network is designed as a fully convolutional network which can process full-size images after the patch-based training. Finally, the overall objective function is:

\begin{equation}
L_{final}=L_{adv}+\lambda L_{pixel}\label{Eq1}
\end{equation}
where $L_{adv}$ is the adversarial loss generated by the pre-trained discriminator and $L_{pixel}$ is the pixel-level loss computed with L1-distance.

\begin{figure}[t!]
	\centering
	\includegraphics[width=0.45\textwidth]{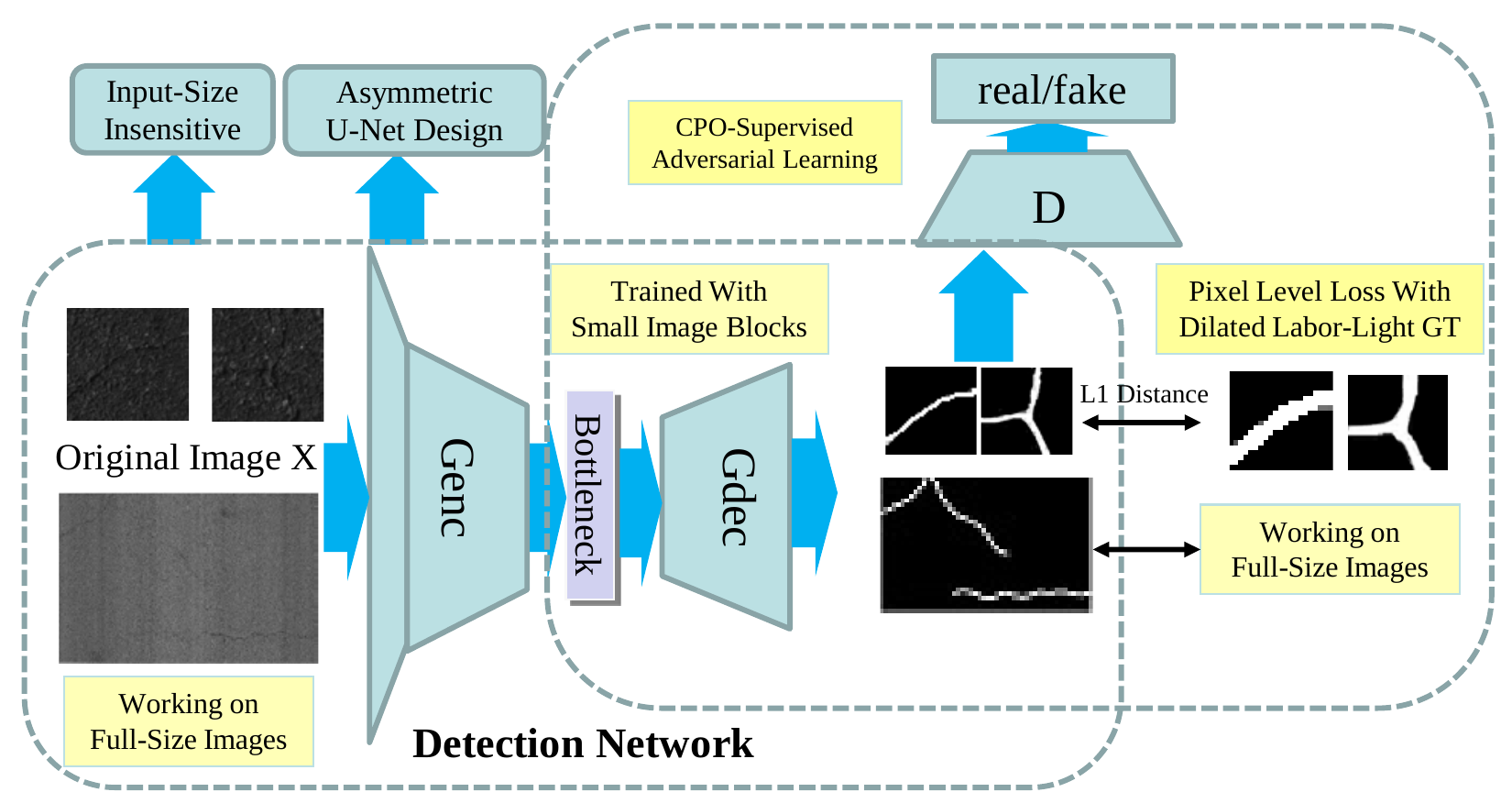}
	\captionsetup{justification=centering}
	\caption{Overview of the proposed method.}
	\label{Fig1}
\end{figure}

\begin{figure}[htbp]
	\centering
	\includegraphics[width=0.50\textwidth]{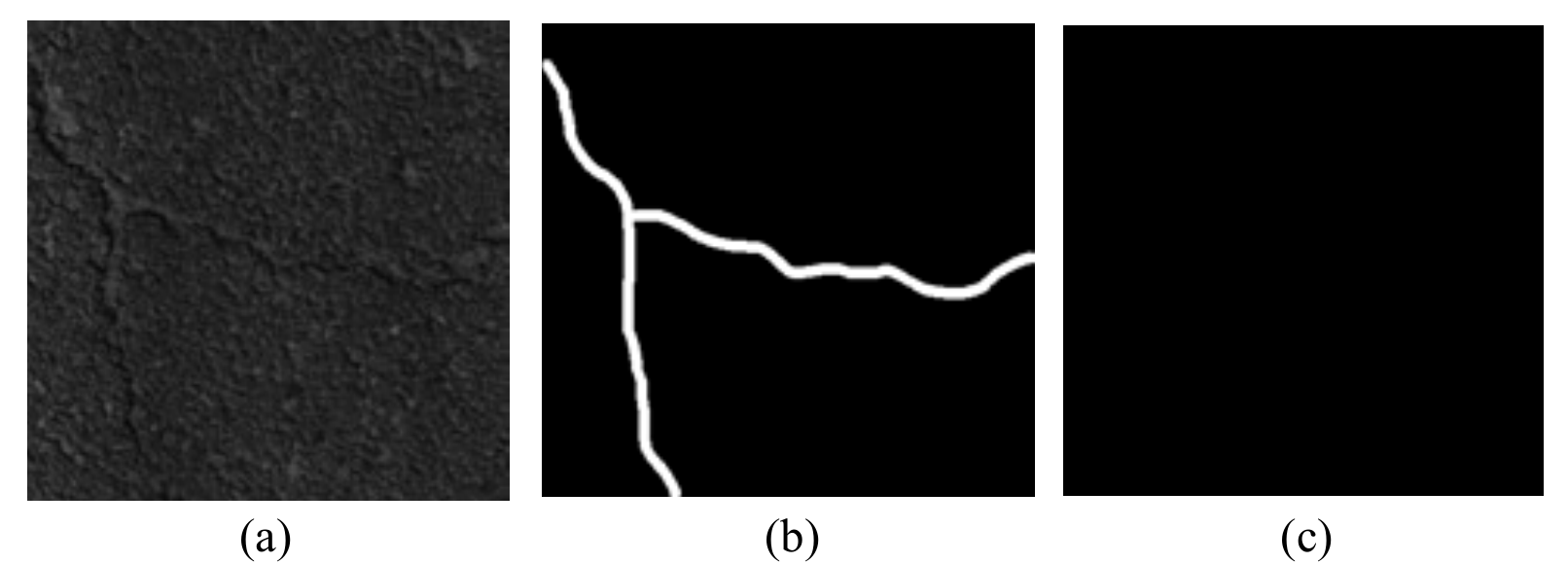}
	\caption{``All Black'' issue encountered when using FCN-based method for pixel-level crack detection: (a) the industrial pavement crack image; (b) the dilated GT-image utilized in the training; and (c) the detection result with the ``well-trained'' U-Net (see Fig. 3).}
	\label{Fig2}
\end{figure}

\begin{figure}[htbp]
	\centering
	\includegraphics[width=0.40\textwidth]{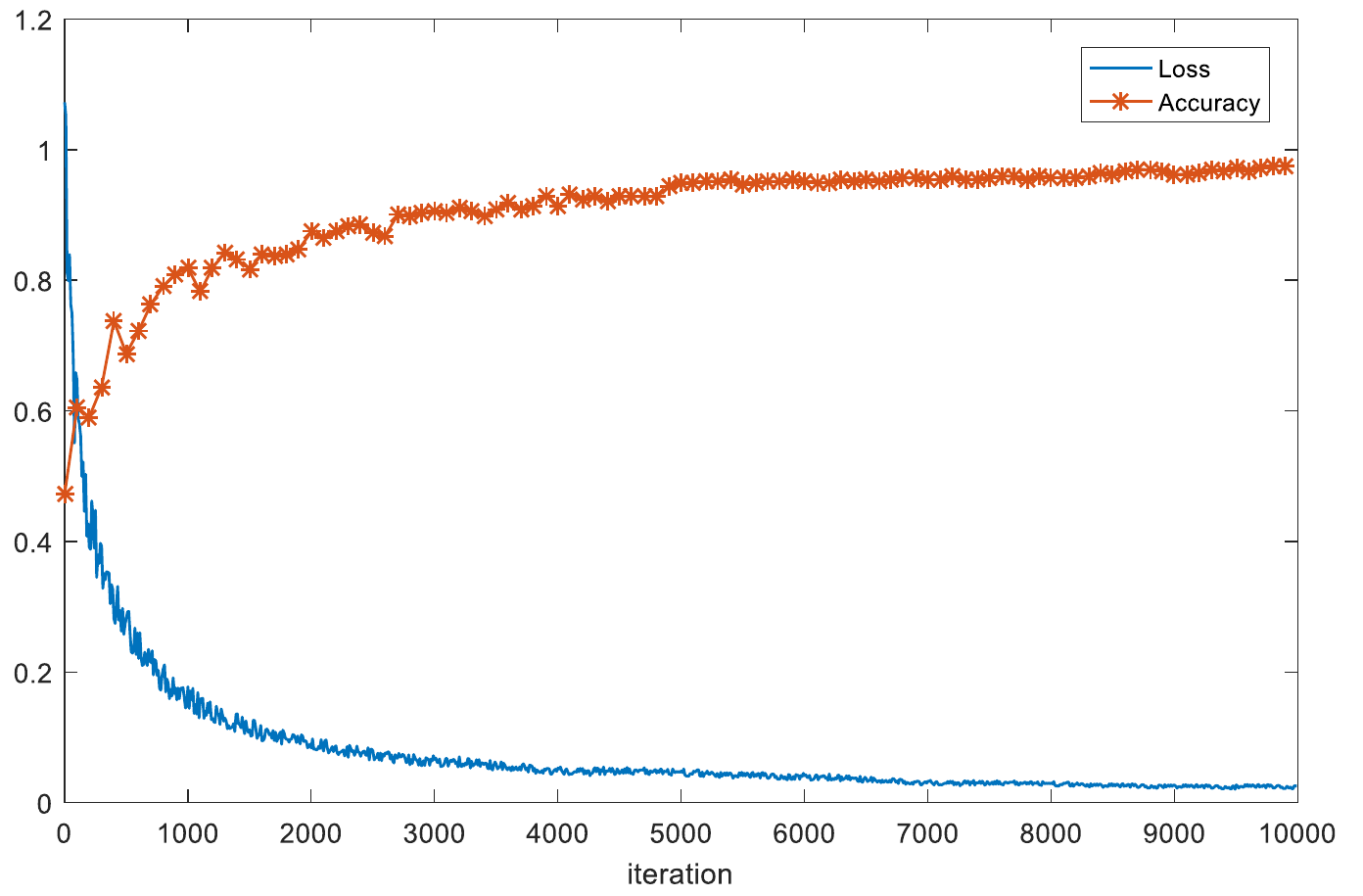}
	\caption{The loss and accuracy curves when training a regular U-Net using industrial pavement images with partially accurate GTs.}
	\label{Fig3}
\end{figure}

\subsection{``All Black'' issue}
This work results from addressing a practical engineering issue that the authors have encountered in industry. At the early attempts, we trained an FCN \cite{rf6} for pixel-level pavement crack detection based on the data and GTs \cite{rf5}. However, the results were not satisfactory: the networks were easily to converge to BG even there were cracks. The most possible reasons were: (1) most cracks in the industrial pavement images were thin and the crack-boundaries were vague, that made it very difficult for per-pixel GT annotation; in practice, the engineers just marked the cracks with 1-pixel curves for simplicity, and they were used as the GTs that were only partially accurate. Such GTs could not match the actual cracks at pixel-level well, which made the loss computation inaccurate, and failed the task. (2) Crack, as a long-narrow object, only occupies a very small area in a full image; and since patch-wise training was equivalent to loss sampling in FCN \cite{rf6}, training an FCN end-to-end with pavement crack images actually worked on extremely imbalanced dataset. Even the network simply classified all the pixels as BG, it still achieved quite a ``good'' accuracy (since BG pixels dominate the whole images), that was the ``All Black'' issue. As shown in Fig. 3, during training, the loss decreases rapidly and approaches to a very low value; however, in Fig. 2, the detection results are all blacks (i.e., all BGs). Moreover, it is worth to mention that other FCN architectures also encounter such problem; here it just takes U-Net as an example. Since the GTs are only partially accurate, existing approaches for solving data imbalance cannot work here.

\subsection{CPO-supervision and one-class discriminator}
Regular FCN-based methods may only produce all-black images as the detection results \cite{rf5, rf7}. In order to address this problem, it adds a new constraint, generative adversarial loss, to regularize the objective function, which will make the network always generate crack-GT detection result; accordingly, the training data are prepared with crack patches only (i.e., CPO-supervision), without involving any non-crack patches and ``all black'' patches. As shown in Fig. 1, the adversarial loss is provided by a one-class discriminator obtained via pre-training the DC-GAN \cite{rf41} only with crack-GT-like patches. It is well-known that the DC-GAN can generate real-like images from random noise by conducting the training with a max-min two-player game, in which a generator is used to generate real-like images and a discriminator is used to distinguish between real and fake images. As verified in \cite{rf38}, it was better for \textit{G} to maximize $log(D(G(z)))$ instead of minimizing $log(1-D(G(z)))$. Therefore, the actual optimization strategy is to optimize the following two objectives alternatively \cite{rf54}:

\begin{equation}
	\begin{aligned}
	\max\limits_{D}\mathop{V}\limits_{D}(D, G) = E_{x\sim p_{d}(x)}[logD(x)]\\+E_{z\sim p_{d}(z)}[log(1-D(G(z)))]\label{Eq2}
	\end{aligned}
\end{equation}

\begin{equation}
\max\limits_{G}\mathop{V}_{G}(D, G)=E_{z\sim p_{d}(z)}[log(D(G(z)))]\label{Eq3}
\end{equation}
where \textit{x} is the image from the real data (crack-GT-like patches) with distribution $p_{d}(x)$; \textit{z} is the noise vector generated randomly from Gaussian distribution $p_{d}(z)$; and \textit{D} is the discriminator and \textit{G} is the generator set up with convolutional and deconvolutional kernels, respectively. In practice, whether a sample is real or fake depends on the data setting. In accordance with the CPO-supervision, only crack-GT patches are contained in the real image-set for training the DC-GAN. With such setting, the discriminator will only recognize crack-GT patch as real and treat all-black patch as fake, which prevents the network to generate all-black (fake) image as the detection result, thus overcoming the ``All Black'' issue. Such discriminator is named one-class discriminator. In the implementation, the crack-GT data are further augmented by manually marking a bunch of ``crack'' curves and sampling the patches accordingly, as indicated in Fig. 4.

\begin{figure}[t!]
	\centering
	\includegraphics[width=0.45\textwidth]{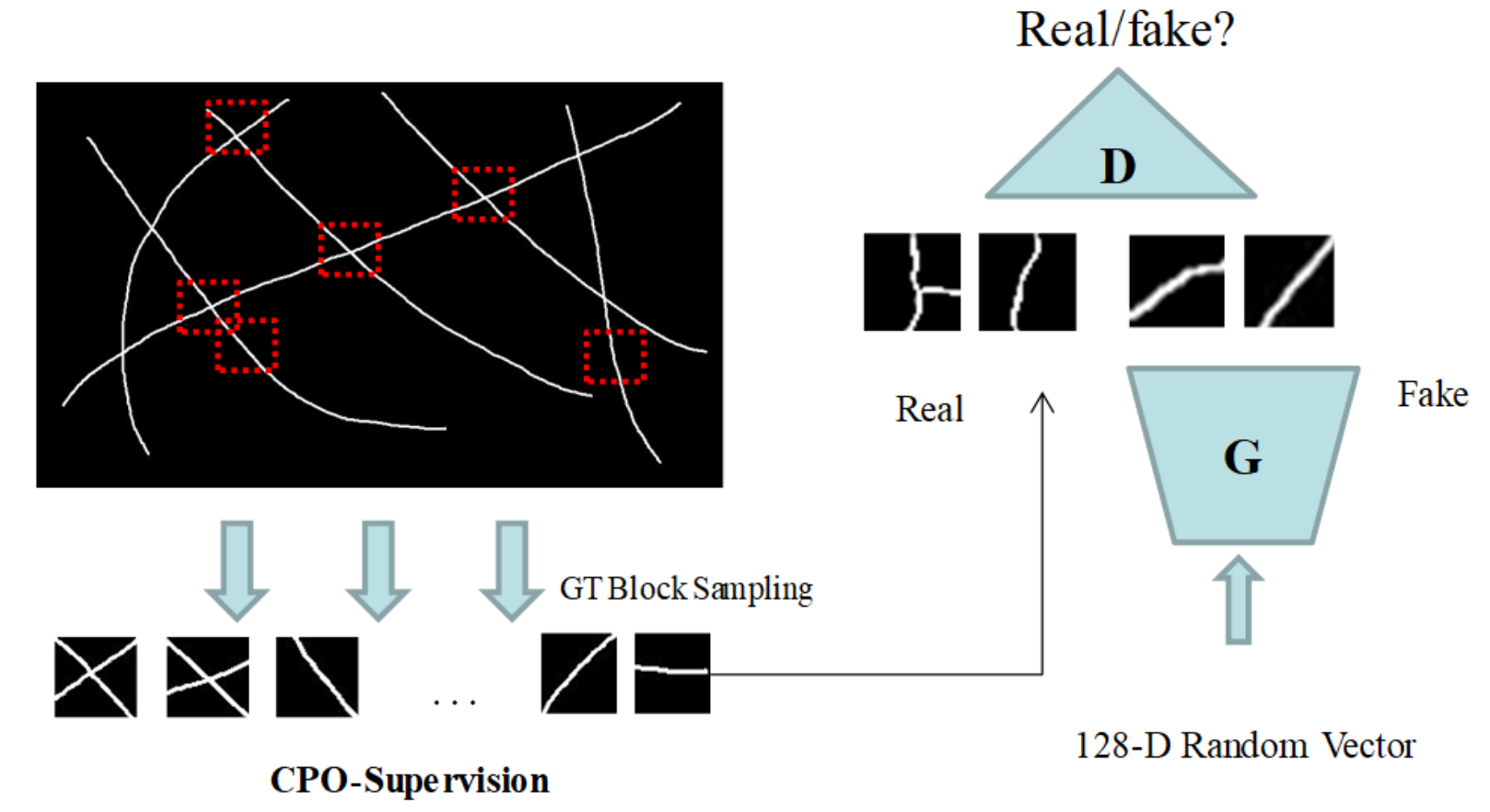}
	\caption{Pre-train a one-class DC-GAN with augmented GTs based on CPO-supervision. The real crack-GT data are augmented with manually marked ``crack'' curves.}
	\label{Fig4}
\end{figure}

In Fig. 5, after training, the discriminator of the well-trained DC-GAN is concatenated to the end of the asymmetric U-Net generator to provide the adversarial loss for end-to-end training. Since the output of the generator serves as a fake image, the adversarial loss is:

\begin{equation}
L_{adv}=-E_{x\in I}[logD(G(x))]\label{Eq4}
\end{equation}
Here, different from pre-training the DC-GAN in Eq. (2), \textit{x} is the crack-patch, and \textit{I} is the training set containing crack patches only; \textit{G} is set up with the asymmetric U-Net architecture illustrated in Fig. 5, and \textit{D} is the pre-trained one-class discriminator. 
%\IEEEpubidadjcol

\begin{figure}[t!]
	\centering
	\includegraphics[width=0.45\textwidth]{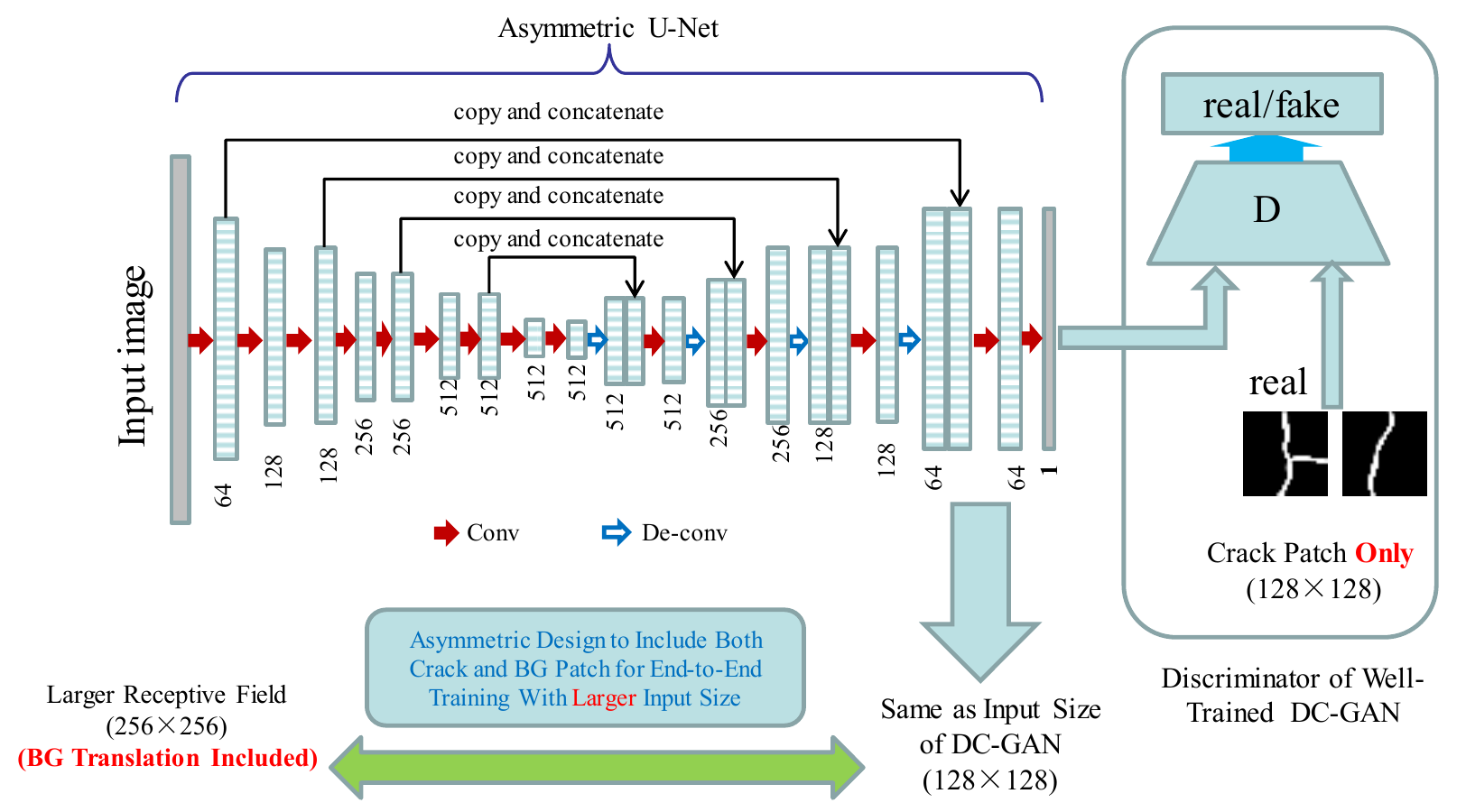}
	\caption{Asymmetric U-Net with larger input image (under larger field of view) with CPO-supervision.}
	\label{Fig5}
\end{figure}

\subsection{Asymmetric U-Net for BG-image translation}
In subsection III-B, it introduced the CPO-supervision and generative adversarial learning to force the network to always generate crack-GT patches and prevent the ``All Black'' phenomenon. However, for a crack detection system, it should be able to process both crack and non-crack/BG pavement images. Normally, the discriminator should treat all-black patch as real to represent the BG-image translation result, such as directly applying the original pix2pix GAN \cite{rf40}; unfortunately, treating all-black patch as real will encourage the network to generate all-black images as the detection results which is against solving the ``All Black'' issue. In order to include the translation of BG-image with CPO-supervision, it replaces the regular U-Net generator in the original pix2pix GAN with the proposed asymmetric U-shape generator which inputs a larger size \textit{crack} patch (256$\times$256) and outputs a smaller \textit{crack-GT} image (128$\times$128) for the end-to-end training. In accordance with the CPO-supervision, the larger input image has to be a crack image so that the \textit{correct} output will always be a crack-GT patch recognized by the discriminator as real. With such setting, the network is able to translate both crack image and BG-image after the training. It is detailed later.

\textit{Receptive field analysis under larger field of view}: To understand how the asymmetric design is able to include BG-image translation ability by only using crack samples for the training, it first performs a receptive field analysis under larger field of view. In Fig. 6, there is a DCNN network, as a classification network, with an $m\times m$ image patch as input, and the output is a single neuron representing the class label of the input image patch. When the same DCNN network is fed a larger size input image, it will output multiple neurons, and each neuron represents a class label of the corresponding image patch of size $m\times m$ ``sampled'' from the larger input image. For example, when the network's input is an image of $m\times 3m$, the output has five neurons (the number of neurons depends on the down-sampling rate of the DCNN) which represent class labels of five image patches including both crack and non-crack samples of size $m\times m$ (from left to right, the first three neurons represent crack-samples and the last two represent BG-samples). Indeed, under the multi-layer convolutional mode, each neuron actually has a receptive field with a specific size; since the convolutional layer is input-size insensitive, operating the network under larger receptive field actually realized a multi-spot image sampling with the image size equal to the receptive field of the neuron \cite{rf5}. Thus, when performing an image translation using a deep convolutional neural network with a larger input image, the process is equal to translating multiple smaller image samples at the same time (the size is equal to the receptive field of the original image translation network).

\begin{figure}[t!]
	\centering
	\includegraphics[width=0.45\textwidth]{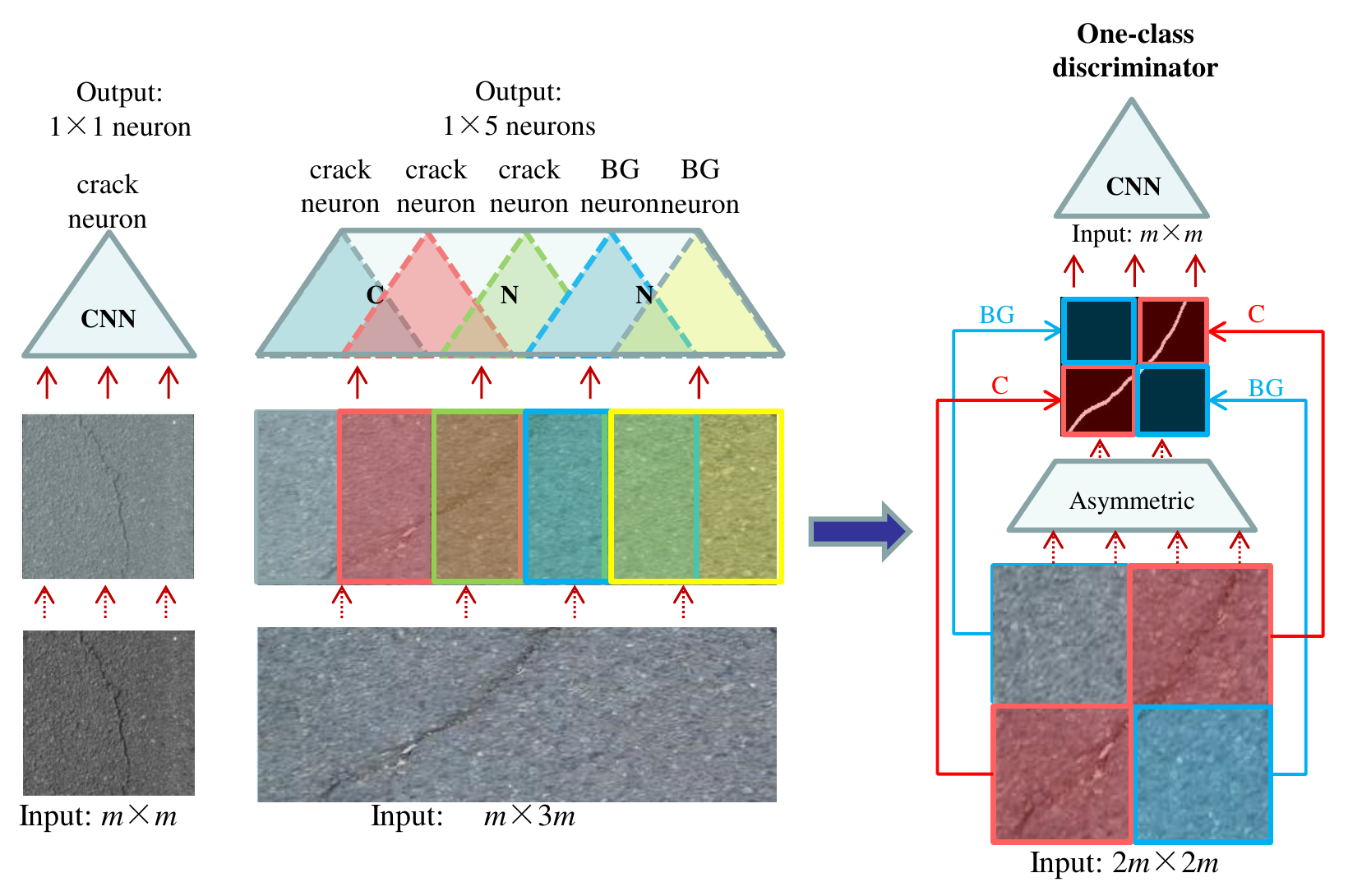}
	\caption{Receptive field analysis under larger field of view: with a larger input image, the CNN realizes multi-spot sampling with the same receptive field. At the right, the asymmetric network architecture includes the image translation of both crack and BG samples; however, the training data contains crack patch only.}
	\label{Fig6}
\end{figure}

According to the analysis, as in Fig. 6, when a crack image with the size larger than the input-size of the discriminator is input to the asymmetric U-Net, and passes through the network; the network will produce a downsampled image patch that exactly \textit{matches the input-size} of the discriminator. The output will be treated as a single image by the one-class discriminator for the generative adversarial learning which still maintains the working mechanism of COP-supervision. However, since the network is trained to translate a larger crack image to a downsampled crack-GT image, it includes the translation of both crack and non-crack image samples inherently. In this way, the network can be trained to process both crack and BG images. Refer to Fig. 6.

\subsection{L1 loss with dilated GTs}It introduces the CPO-supervised generative adversarial learning and the asymmetric U-Net to prevent the ``All Black'' phenomenon; however, it is only an image-level supervision that does not specify the exact location of the cracks in the generated image. As analyzed before, one of the reasons for the ``All Black'' issue is the pixel-level mismatching due to the inaccurate GTs. Thus, it introduces the dilated-GT to specify a relatively larger crack area to ensure that it covers the actual crack locations, and if a detected crack pixel is in the dilated area, it is treated as a true positive. The experiments demonstrate that by combining the CPO-supervised adversarial loss and the loosely-supervised L1 loss, the network can be trained to generate cracks in the expected locations. Following \cite{rf4}, it marks the cracks with 1-pixel-width curves, and crops crack patches and partially accurate crack-GT patches from the original pavement images and the images with partially accurate GT, respectively. Then the partially accurate 1-pixel-width GTs were dilated three times using a disk structure with radius of 3 to generate the dilated GTs which are used to provide the loosely supervised pixel-level loss:

\begin{equation}
L_{pixel}=-E_{x\in I, y}[\|y-G(x)\|_{1}]\label{Eq5}
\end{equation}
where \textit{x} is the input crack patch; \textit{y} is the dilated GT; \textit{I} is the dataset of larger size crack patch (256$\times$256 comparing with the output size of 128$\times$128) used for end-to-end training; \textit{G} is the asymmetric U-Net;  and  \textit{D}  is  the discriminator.

Overall, the final objective function is:

\begin{equation}
L_{final}=L_{adv}+\lambda L_{pixel}\label{Eq6}
\end{equation}
The pixel-level loss is normalized during training and $\lambda=0.30$ is determined via grid search with step size 0.05. Fig. 7 shows the detection result of a sample image. Moreover, once the training is finished, the asymmetric U-Net generator itself will serve as the detection network to translate the original pavement image to the result image.

\begin{figure}[t!]
	\centering
	\includegraphics[width=0.46\textwidth]{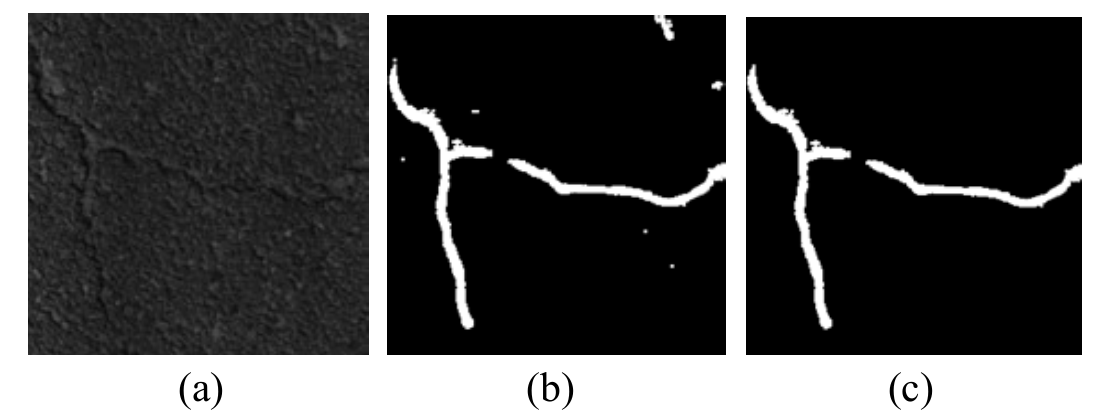}
	\caption{Detection results of CrackGAN: (a) industrial pavement image suffered from ``All Black'' issue; (b) detection result of CrackGAN; (c) the final result image after removing the isolated noises.}
	\label{Fig7}
\end{figure}

\subsection{Working on full-size images}
Notice that the network is trained with small image patches; however, under industry settings, the image size is much larger (2048$\times$4096 pixels). A traditional solution to process large input image is to sample it into smaller image patches from the full-size image and do the processing patch-by-patch, named window-sliding strategy \cite{rf29, rf4}; however, it is very inefficient \cite{rf5}. In our approach, the asymmetric U-Net is designed as a fully convolutional network, and it can work on images of arbitrary sizes seamlessly. In addition, such fully convolutional processing mechanism is quite efficient, which does not involve redundant convolutions as discussed in \cite{rf5}.
\subsection{Implementation details}
\textit{Network architecture}: Fig. 5 presents the architecture of the asymmetric U-Net. The first layer is configured with 7$\times$7 convolutional kernels with stride 2 and is followed by a rectified linear unit (ReLU) \cite{rf53}; and it serves as the asymmetric part of the U-Net generator, which realize a 2-time downsampling of the larger input images and output the feature maps with the same size as the final output of the asymmetric U-Net. Then the remaining layers of the encoding and decoding parts are following the regular U-Net architecture \cite{rf49}. The encoding part consists of four repeated convolutional layers with 3$\times$3 kernels and the stride is 2; and each convolutional layer is followed by a ReLU layer. After each of the first three convolutional layers, the number of convolutional channels is doubled. The decoding part consists of four 3$\times$3 de-convolutional layers that up-samples the feature maps; the input of each de-convolutional layer is the output of the last layer concatenated with the corresponding feature map from the encoding part, then followed by a regular convolutional layer. After the last de-convolutional layer, another regular convolutional layer with Tanh activation \cite{rf47} is utilized to translate the 64-channel feature map to the 1-channel image, and it is compared with the dilated-GT for L1 loss computation according to Eq. (5). In summary, the network architecture is as follows. The encoding part:\\
C\_64\_7\_2 - ReLU - C\_128\_3\_1 - ReLU - C\_128\_3\_2 - ReLU - C\_256\_3\_1 - ReLU - C\_256\_3\_2 - ReLU - C\_512\_3\_1 - ReLU - C\_512\_3\_2 - ReLU - C\_512\_3\_1 - ReLU - C\_512\_3\_2 - ReLU\\
The decoding part:\\
DC\_512\_3\_2 - ReLU - C\_512\_3\_1 - ReLU - DC\_256\_3\_2 - ReLU - C\_256\_3\_1 - ReLU - DC\_128\_3\_2 - ReLU - C\_128\_3\_1 - ReLU - DC\_64\_3\_1 - ReLU - C\_64\_3\_1 - ReLU - C\_1\_3\_1 - Tanh\\
Here, the naming rule follows the format: ``layer type\_channel number\_kernel size\_stride''. ``C'' denotes convolution; ``DC'' is de-convolution; and Tanh is the Tanh activation. For instance, ``C\_64\_7\_2'' means that the first layer is a convolutional layer and the number of channels is 64, the kernel size is 7 and the stride is 2.

\textit{Network training}: The training is a two-stage strategy which employs transfer learning at two places, the one-class discriminator and the encoding part of the generator. First, the DC-GAN is trained with the crack-GT patches of 128$\times$128-pixel as described in subsection III-B, aiming at training a discriminator with strong crack-pattern recognition ability to provide the adversarial loss for the end-to-end training at the second stage. A total of 60,000 dilated crack-GT patches with various crack patterns are used. The other training settings follow \cite{rf41}: the Adam optimizer \cite{rf55} is used, the learning rate is 0.0002, the parameters for momentum updating are 0.9, the batch size is 128 and the input ``noise'' vector is 128 dimensions. A total of 100 epochs (each epoch is \textit{total images/batch size} = 60000/128 iterations) are run to obtain the final model. Then the well-trained discriminator is concatenated to the end of the asymmetric U-Net to provide the adversarial loss at the second stage. Refer Fig. 5 and Eq. (4). 

\begin{figure}[t!]
	\centering
	\includegraphics[width=0.45\textwidth]{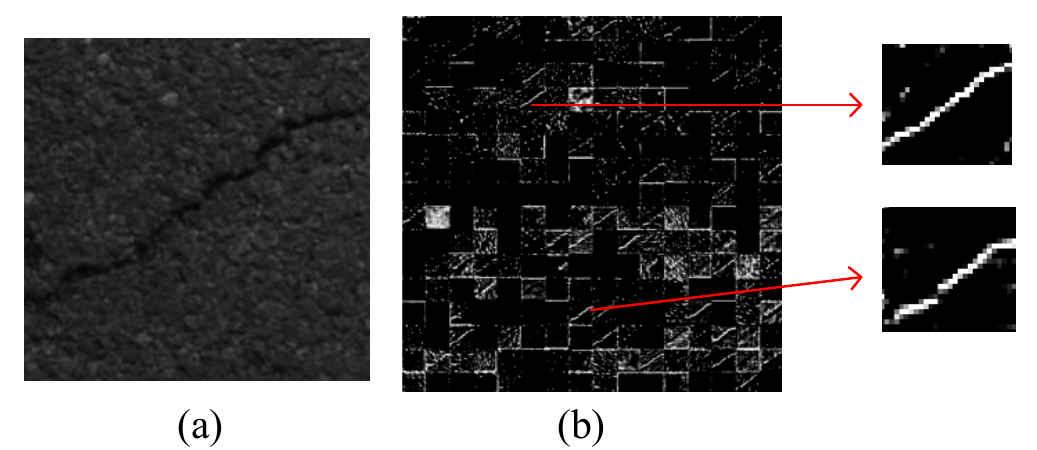}
	\caption{Weakly supervised learning is able to learn rich crack pattern information: (a) the image patches inputs into the classification network; (b) the feature maps after the first convolutional layer.}
	\label{Fig8}
\end{figure}

Inspired by \cite{rf4}, it also pre-trains the encoding part of the generator under the classification setting. Zhang et al. \cite{rf4} showed that by performing an image-block classification task, the network was able to extract the relevant crack patterns; and the learned knowledge could be transferred to ease the training of an end-to-end detection network \cite{rf5}. Fig. 8 is the low-level feature maps of a classification network trained with crack and non-crack patches \cite{rf4}. The classification network is configured by adding a fully connected layer at the end of the encoding part (bottleneck) of the asymmetric U-Net and the output dimension is 2 representing crack and non-crack with labels 0 and 1. The training samples are crack and non-crack patches of 256$\times$256. It shows that the network extracted same crack pattern as the original image, i.e., the network is able to learn useful information with the weakly supervised information, crack/non-crack image labels only. Then the well-trained parameters are used to initialize the encoding part of the generator for the end-to-end training; and the other settings are same with the DC-GAN except replacing the generator with the asymmetric U-Net and changing the objective function according to Eq. (6).

\begin{figure}[t!]
	\centering
	\includegraphics[width=0.40\textwidth]{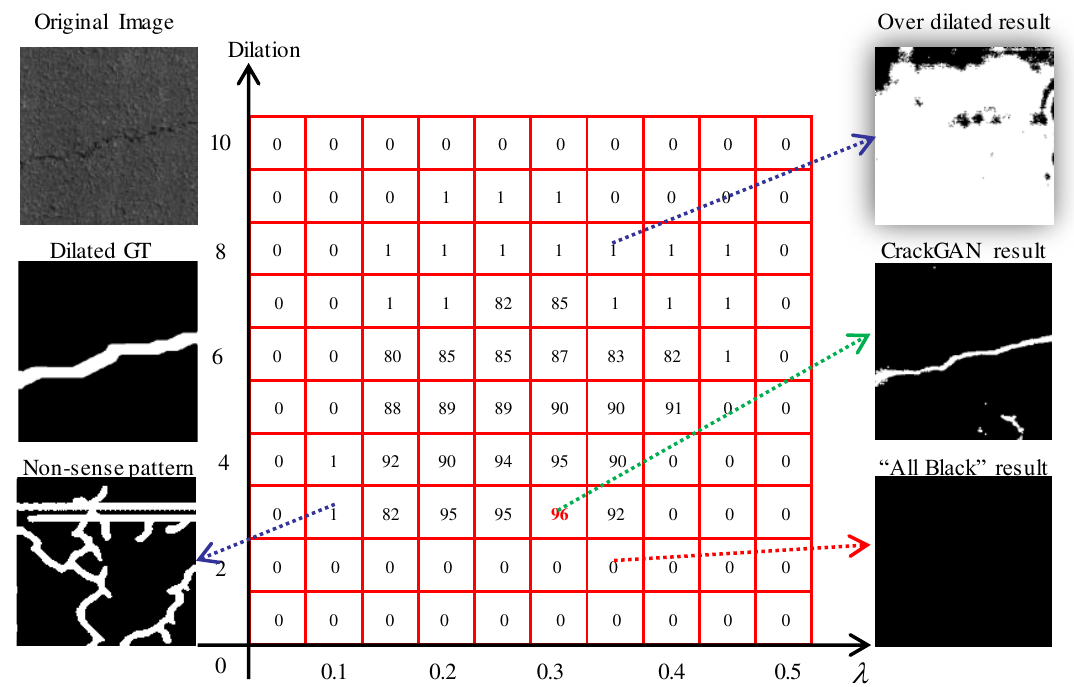}
	\caption{Grid search board with HD-scores utilized to determinate the optimal parameters $\lambda$ and dilation-scale. The optimal parameters are determined according to the best HD-score. Testing results of the three main failure cases are also present in the figure including the ``All Black'' result, over-dilated result, and the non-sense patterns which overlooked the pixel-level loss.}
	\label{Fig9}
\end{figure}

\textit{Parameter selection}: The parameters, $\lambda$ in Eq. (6) and the dilation scale, are determined by grid search. From the grid search board in Fig. 9, the dilation scale (number of dilation times with the disk structure) should be between 3 and 7, and the $\lambda$ should be between 0.15 and 0.4 for an effective detection. Dilation scale less than 3 will cause the ``All Black'' issue due to the pixel-level mismatching and the data imbalance, while too big dilation ($>$8) could not provide meaningful crack location information and would produce useless output. In addition, a larger $\lambda$ tends to fail the task, but a small $\lambda$ (0.15 $<\lambda<$0.4) can work well, which indicates that the adversarial loss is very important to succeed the training under the industrial setting. It can be observed from the grid board that the HD-score is either a good one ($>$80) or a very small one (0 or 1) which indeed represents the two different model statuses, well-trained or failed. However, the causes of the failures could be grouped into three main cases: over-dilated, ``All Black'' problem, or the useless output pattern which overlooks the pixel-level loss with a small $\lambda$, as indicated in Fig. 9. 

\begin{figure}[htbp]
	\centering
	\includegraphics[width=0.50\textwidth]{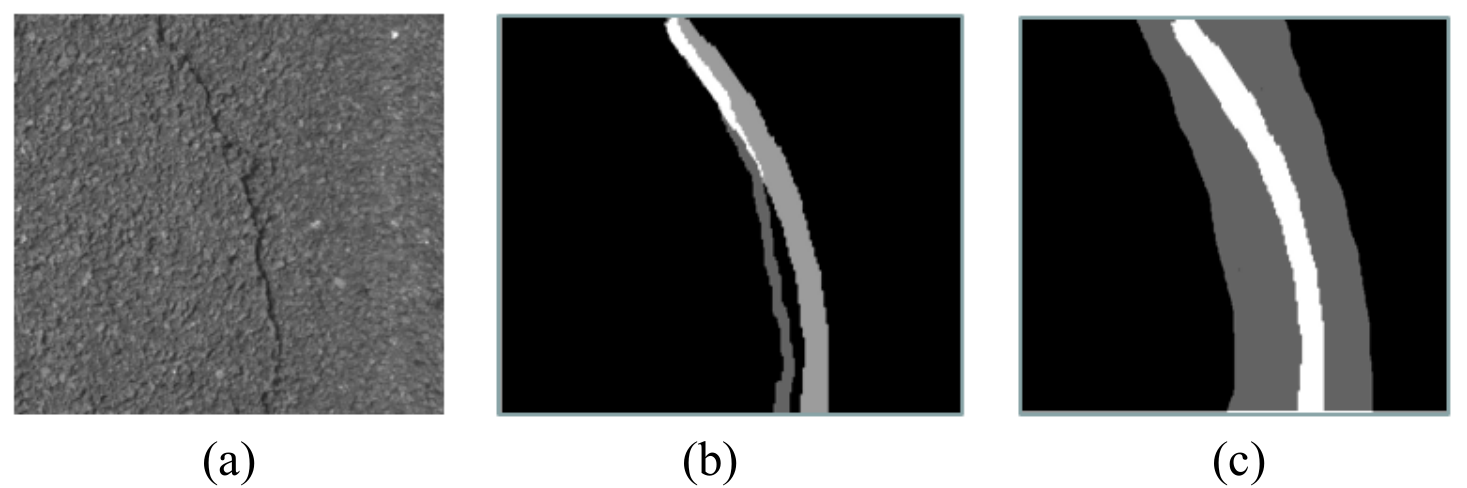}
	\caption{Illustration of pixel-level mismatching: (a) a crack image; (b) detection result overlapped with GT-image dilated once using a disk structure with radius 3; and (c) detection result overlapped with GT-image dilated four times. The transparent areas represent the dilated GT-cracks with different dilation scales and the green areas represent the detected cracks.}
	\label{Fig10}
\end{figure}

\begin{figure}[htbp]
	\centering
	\includegraphics[width=0.40\textwidth]{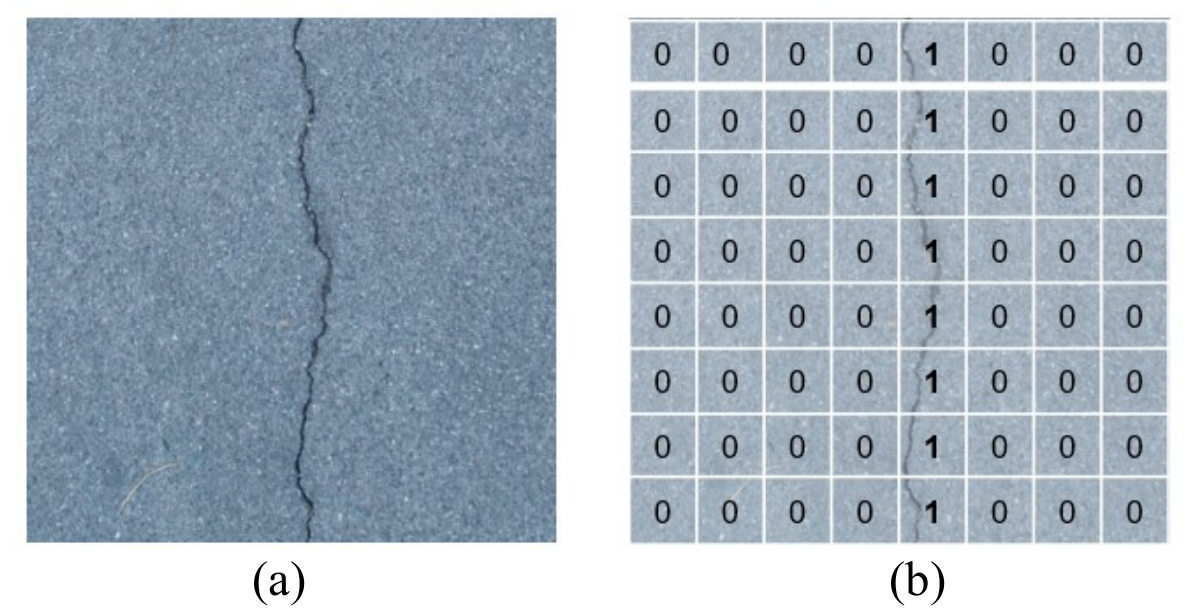}
	\caption{Region-based evaluation: (a) the original crack image; (b) illustration of counting the crack and non-crack regions. The squares with label ``1s'' are the crack regions, and with label ``0s'' are BG-regions.}
	\label{Fig11}
\end{figure}

\section{Experiments}

\begin{figure*}[htbp]
	\centering
	\includegraphics[width=0.9\textwidth]{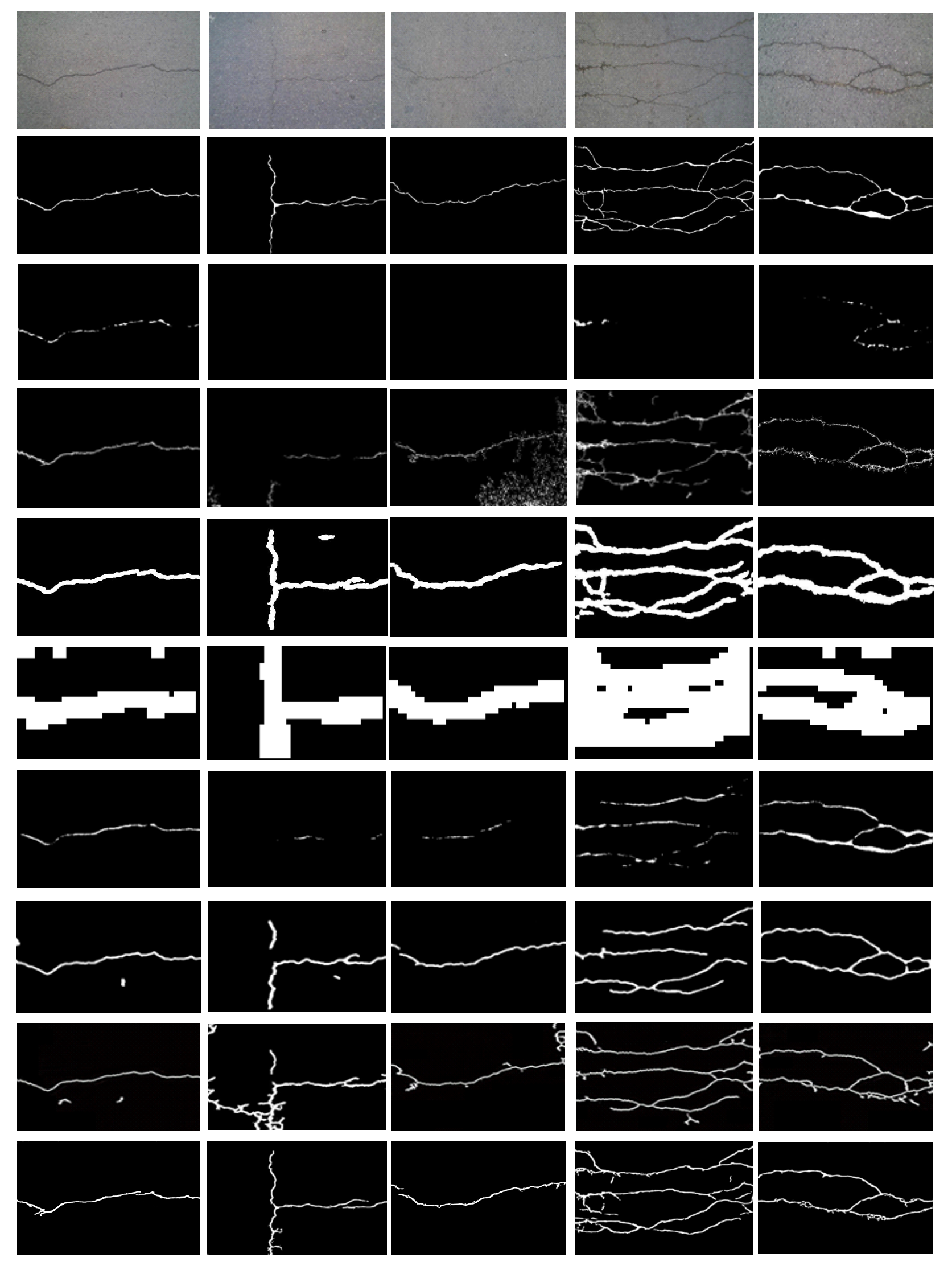}
	\caption{Comparison of the detection results on CFD using different methods. From top to bottom are: original images, GT images, results of CrackIT, results of MFCD, results of CrackForest, results of \cite{rf29}, results of FCN-VGG, results of DeepCrack, results of Pix2pix GAN, and results of CrackGAN, respectively.}
	\label{Fig12}
\end{figure*}

\subsection{Dataset and Metrics}
CFD \cite{rf3}, the CrackGAN dataset (CGD) collected by the authors, and dataset \cite{rf17} are utilized for evaluation. CFD contains 118 pavement crack images (480$\times$320-pixel each) obtained by people standing on the road using an iPhone, the ground truths are carefully marked at pixel level which is labor-intensive. The image quality is high and the background is smooth and clean. CGD is a dataset with 400 pavement crack images (2048$\times$4096-pixel each) collected by the authors using a line-scan industrial camera mounted on the top of a vehicle running at 100km/h; and the camera scans 4.096-meter width road surface and produces a pavement image of 2048$\times$4096-pixel for every 2048 line-scans (i.e., 1 pixel represents 1$\times$1 mm$_{2}$ area). Most of the cracks are thin, and sometimes even hard to be recognized by human. Furthermore, it is infeasible to obtain the accurate GTs at pixel-level; thus, the cracks are represented by 1-pixel curves roughly marked by the engineers in a labor-light way, and it is named partially accurate GTs). However, such GTs may not match the true crack locations accurately, and processing them is much more challenging. The proposed algorithm can achieve the best results using both ``accurate'' and ``partially accurate'' datasets that demonstrate its robustness as well. For CFD and CGD, the data are augmented following \cite{rf4} to facilitate the training; and the training-test ratio is 2:1. Dataset \cite{rf17} has industrial images from five different capture systems: Aigle-RN has 38 images with annotation, ESAR has 15 images with annotations, and LCMS has 5, LRIS has 3 and Tempest has 7 images with annotations, respectively. The GTs are marked at pixel level. To our best knowledge, it is the only public pavement crack dataset from industry; and it contains relatively few images, they are used for testing only.

Different from most object detection tasks \cite{rf56}, the intersection over union (IOU) is not suitable for evaluating crack detection algorithms \cite{rf57}. As shown in Fig. 10, crack, as a long-narrow target, only occupies a very small area, and the image consists mainly of BG pixels \cite{rf57}. With the fact that the precise pixel-level GTs are difficult to obtain, it is impossible to obtain the accurate intersection area. As shown in Fig. 10 (b) and Fig. 10 (c), it is obvious that the detection results are very good; however, the IOU values are very low, 0.13 and 0.2, respectively. According to \cite{rf57}, it employs Hausdorff distance to evaluate the crack localization accuracy. For two sets of points \textit{A} and \textit{B}, the Hausdorff distance can be calculated with:

\begin{equation}
H(\mathrm{A,B})=max[h(\mathrm{A},\mathrm{B}),h(\mathrm{A,B})]\label{Eq7}
\end{equation} where

\begin{equation}
h(\mathrm{A,B})=max_{a\in \mathrm{A}} min_{b\in \mathrm{B}} \|a-b\|\label{Eq8}
\end{equation}The penalty is defined as:

\begin{equation}
h_{p}\left(\mathrm{A,B}\right)=1/(|\mathrm{A}|)\sum_{a\in \mathrm{A}}sat_{u}min_{b\in \mathrm{B}} \|a-b\|\label{Eq9}
\end{equation}
Here, parameter \textit{u} is the upper limit of the saturation function \textit{sat} which is used to directly get rid of the false positives that are far away from the GTs. Instead of setting \textit{u} as 1/5 of the image width \cite{rf57}, it is set as 50-pixel to emphasize the localization accuracy by eliminating the influence of possible noises from the large BG areas. \textit{A} is the detected crack set and \textit{B} is the GT set, the overall score is:

\begin{equation}
score_{BH}(A, B)=100-\frac{BH(A,B)}{u}\times 100\label{Eq10}
\end{equation}
where

\begin{equation}
BH(A, B)= max[h_{p}(A,B), h_{p}(B,A)]\label{Eq11}
\end{equation}
The Hausdorff distance score (HD-score) can reflect the overall crack localization accuracy, and it is insensitive to the foreground-background imbalance inherent in long-narrow object detection.

In addition, the region-based precision rate (p-rate) and recall rate (r-rate) are used for evaluation, which can measure the false-detection severity and the missed-detection severity, respectively. In Fig. 11, a pavement image of 400$\times$400-pixel is divided into small image patches (50$\times$50-pixel); if there is a crack detected in a patch, marked as ``1s'', it is positive. In the same way, for GT images, if there is a marked curve in a patch, it is a crack patch. Then the region based true positive (TP), false positive (FP) and false negative (FN) can be obtained by counting the corresponding squares, and further be used to calculate the region based precision and recall rates:

\begin{equation}
P_{region}=\frac{TP_{region}}{TP_{region}+FP_{region}}\label{Eq12}
\end{equation}

\begin{equation}
R_{region}=\frac{TP_{region}}{TP_{region}+FN_{region}}\label{Eq13}
\end{equation}
Then the region-based F1 score can be computed as:

\begin{equation}
F1_{region}=\frac{2*P_{region}*R_{region}}{P_{region}+R_{region}}\label{Eq14}
\end{equation}

\begin{table}[htbp]
	\centering
	\begin{center}
		\caption{Quantitative evaluations on CFD}
	\end{center}
	\label{tab:Eval-CFD}
	\begin{tabular}{|c|c|c|c|c|}
		\hline
		Methods & $P_{region}$ & $R_{region}$ & $F1_{region}$ & \textit{HD-score} \\ \hline
		CrackIT & \textbf{88.05}\% & 45.11\% & 59.65\% & 21 \\ \hline
		MFCD & 80.90\% & 87.47\% & 84.05\% & 85 \\ \hline
		CrackForest & 85.31\% & 90.22\% & 87.69\% & 88 \\ \hline
		\cite{rf29} & 68.97\% & \textbf{98.21}\% & 81.03\% & 70 \\ \hline
		FCN-VGG \cite{rf7} & 86.01\% & 92.30\% & 89.04\% & 88 \\ \hline
		DeepCrack & 88.03\% & 94.11\% & 90.96\% & 94 \\ \hline
		Pix2pix GAN & 88.01\% & 90.02\% & 89.01\% & 90 \\ \hline
		CrackGAN & 88.03\% & 96.11\% & \textbf{91.89}\% & \textbf{96} \\ \hline
	\end{tabular}
\end{table}

\begin{table}[htbp]
	\centering
	\begin{center}
		\caption{Quantitative evaluation on CGD}
	\end{center}
	\label{tab:Eval-CGD}
	\begin{tabular}{|c|c|c|c|c|}
		\hline
		Methods & $P_{region}$ & $R_{region}$ & $F1_{region}$ & \textit{HD-score} \\ \hline
		CrackIT & \textbf{89.10}\% & 2.52\% & 4.90\% & 9 \\ \hline
		CrackForest & 31.01\% & 98.01\% & 47.22\% & 63 \\ \hline
		\cite{rf29} & 69.20\% & \textbf{98.30}\% & 81.22\% & 64 \\ \hline
		FCN-VGG \cite{rf7} & 0.00\% & 0.00\% & N/A & N/A \\ \hline
		DeepCrack-1 & 37.01\% & 97.01\% & 53.57 & 65 \\ \hline
		DeepCrack-2 & 0.00\% & 0.00\% & N/A & N/A \\ \hline
		Pix2pix GAN & 0.00\% & 0.00\% & N/A & N/A \\ \hline
		CrackGAN & 87.01\% & 96.01\% & \textbf{91.28}\% & \textbf{96} \\ \hline
	\end{tabular}
\end{table}

\subsection{Overall Performance}
The comparisons are performed on CFD \cite{rf3}, CGD, and dataset \cite{rf17} to justify the state-of-the-art performance.

%\begin{figure}[t!]
%	\centering
%	\includegraphics[width=0.5\textwidth]{Figs/Fig13}
%	\caption{Precision-recall curves. Figures (a), (b), and (c) are the PR-curves on CFD, CGD, and dataset \cite{rf17}, respectively.}
%	\label{Fig13}
%\end{figure}

\begin{figure*}[htbp]
	\centering
	\includegraphics[width=0.90\textwidth]{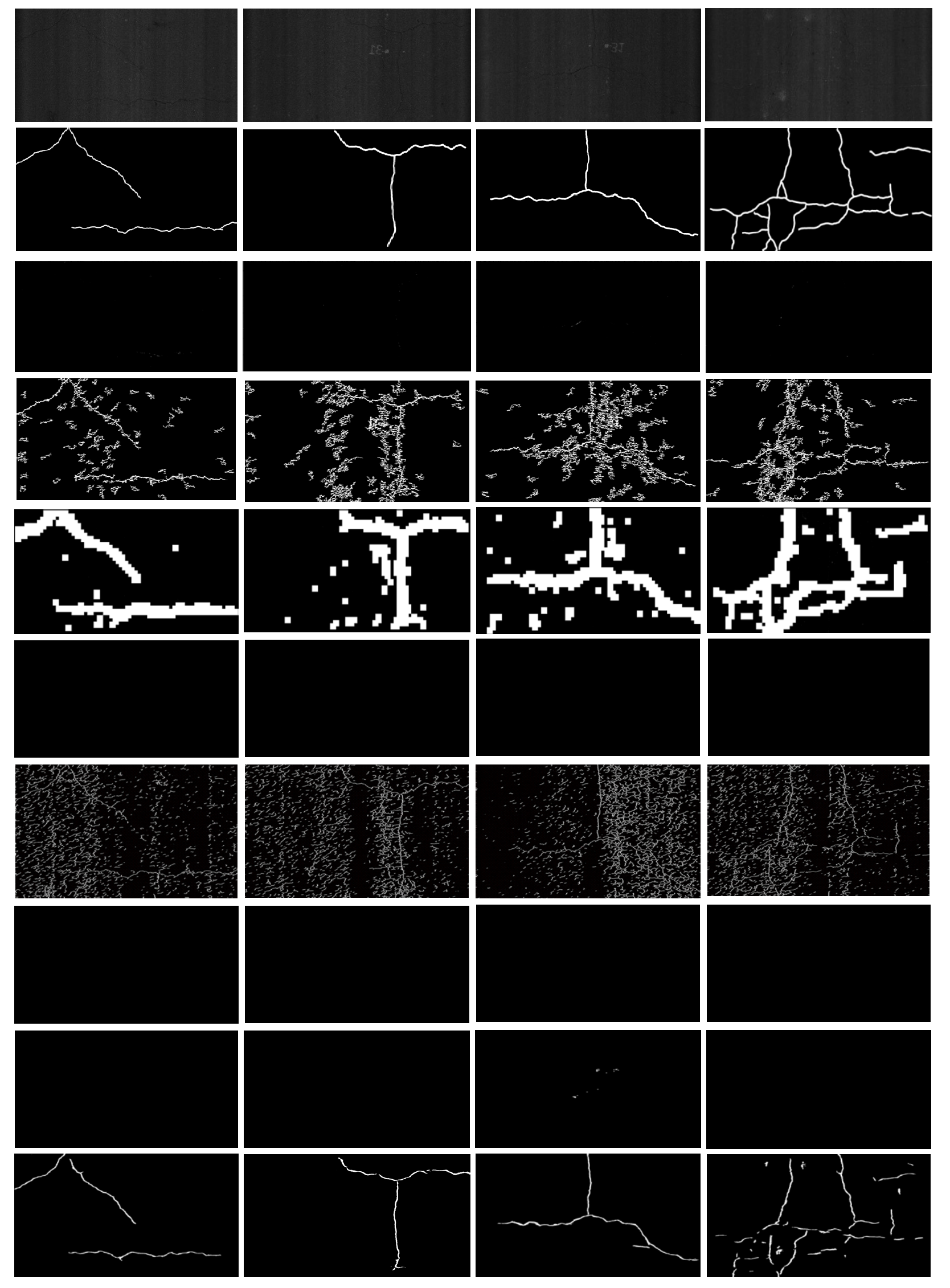}
	\caption{Comparison of the detection results on CGD. From top to bottom are: original images, GT images, results of CrackIT, results of CrackForest, results of \cite{rf29}, results of FCN-VGG, results of DeepCrack-1, results of DeepCrack-2, results of Pix2pix GAN, and results of CrackGAN, respectively.}
	\label{Fig13}
\end{figure*}

\begin{figure*}[htbp]
	\centering
	\includegraphics[width=0.95\textwidth]{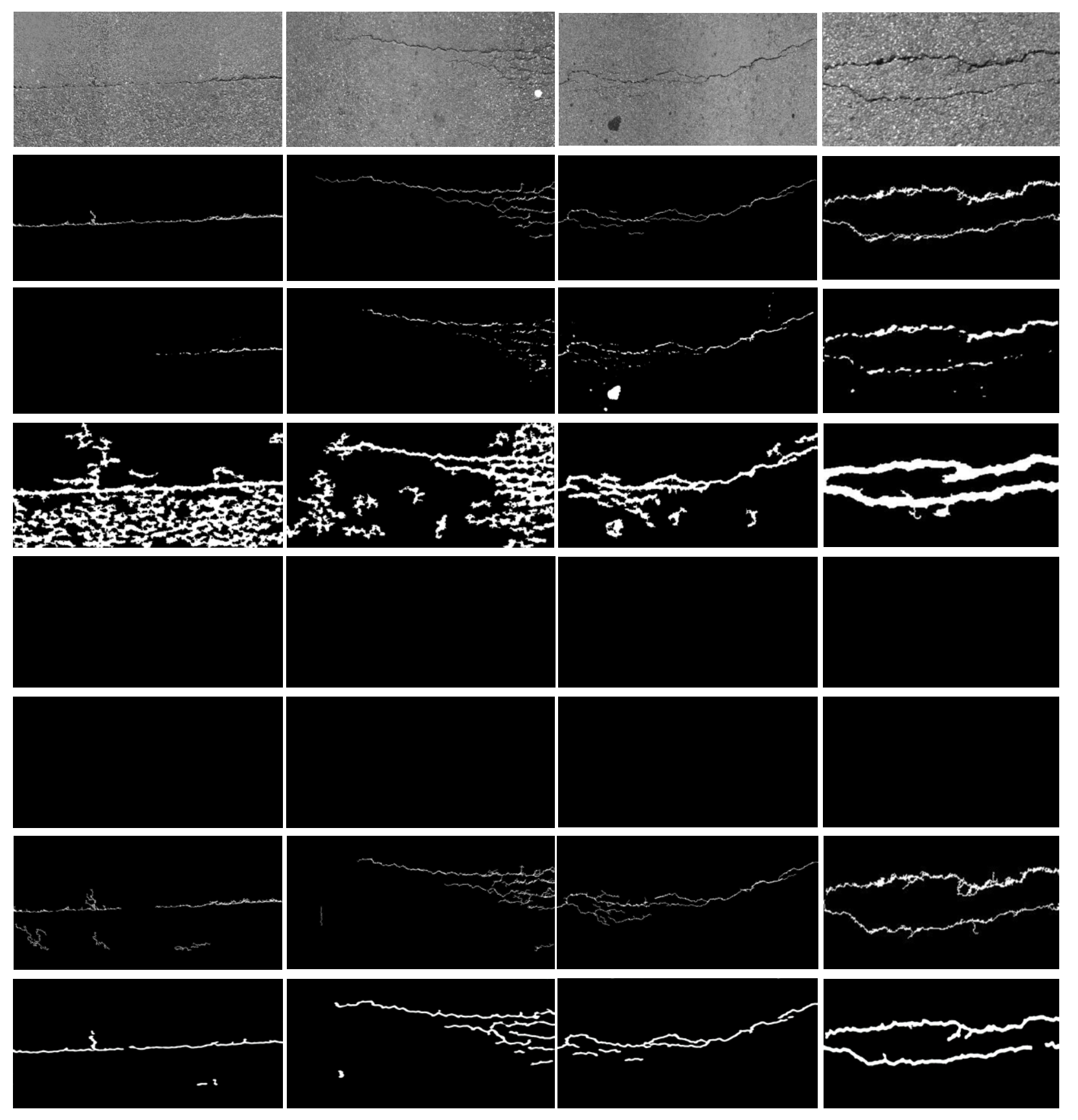}
	\caption{Comparison of the detection results on dataset \cite{rf17}. From top to bottom are: original images, GT images, results of CrackIT, results of CrackForest, results of FCN-VGG, results of Pix2pix GAN, results of MPS \cite{rf17}, and results of CrackGAN, respectively}
	\label{Fig14}
\end{figure*}

\textbf{CFD}: The proposed method is compared with CrackIT-v1 \cite{rf58}, MFCD \cite{rf59}, CrackForest \cite{rf3}, \cite{rf29}, FCN-VGG \cite{rf7}, Pix2pix GAN (with U-Net as the generator) \cite{rf40}, and DeepCrack \cite{rf36} on CFD; and the related results are shown in Fig. 12 and Table I. CrackIT introduced the traditional mean and standard deviation (STD) for crack patch selection, and utilized some post-processing for pixel-based crack detection. However, the features with mean and STD are not able to select the crack patches well, especially when the cracks are thin; thus, the false negative rate is high, and it cannot even detect any cracks in the second and third sample images from Fig. 12. MFCD developed a complex path verification algorithm to link candidate crack seeds for the detection; however, it might also connect the false positives and generate fake cracks. As shown in Fig. 12, it produces many noises in the third image with non-smooth background. CrackForest employed integral channel information with 3 colors, 2 magnitudes and 8 orientations for feature extraction and applied random forest for crack token mapping; and the histogram difference between crack and non-crack regions was used for noise removal. As shown in Fig. 12, it achieves very good results on the images whose backgrounds are smooth and clean. However, the performance deteriorates when processing the industrial images as shown in Figs. 13 and 14. \cite{rf29} was a patch-level crack detection method which trained a deep classification network for crack and non-crack patch classification; it could not provide accurate crack locations as shown in Fig. 12. FCN-VGG was a pixel-level crack detection method of which the accurate pixel-level GTs were needed to train the FCN-based network end-to-end. Similar to the results reported in the original papers, it failed when detecting thin cracks. DeepCrack achieves very good results on CFD due to the multi-scale hierarchical fusion; however, the training relied on accurate GTs and the method would fail easily when the GTs are biased. Pix2pix GAN \cite{rf40} was an image-to-image translation network with U-Net as the generator which introduced generative adversarial learning for image style translation originally. However, as discussed in section III-C, the discriminator would treat both crack and non-crack as real which immediately would weaken the crack-patch generation ability, that makes the network similar to the regular U-Net; therefore, it achieves similar results as FCN-based methods, and also encounters ``All Black'' problem as shown in Figs. 13 and 14. CrackGAN introduces CPO-supervision and the asymmetric U-Net architecture to build the one-class discriminator for generative adversarial learning, which enhances the crack patch discrimination ability by treating the all-black patch as fake images to avoid the data imbalance problem inherent in crack-like object detection, and finally improves the crack detection ability, especially for thin and tiny crack detection.  As shown in Fig. 12 and Table I, it achieves the best results.

It is worth to mention that in Tables I, II, and III, some p-rates and r-rates of the CrackGAN are not the maximum values, but they do not affect the state-of-the-art performance. For example in Table I, CrackIT achieved best p-rate (88.05\%) even it missed quite a lot of cracks; because the precision is calculated with TP/(TP+FP), if FP is small; even FN is very large, the p-rate can still be large. Similarly, \cite{rf29} achieved very good r-rate (98.21\%) even the patch level detection will cause a lot of false positives, because the recall rate is calculated with TP/(TP+FN) which does not take into account the FP. Therefore, only p-rate or r-rate cannot represent the performance of state-of-the-art crack detection algorithm. Refer Table I for the quantitative results.

\textbf{CGD}: The related results on CGD are shown in Fig. 13 and Table II. Similar to the results on CFD, CrackIT misses most cracks and MFCD introduces many noises because of the thin cracks and textured background. CrackForest introduces many noises, among which quite a lot of them connected to the true crack regions; and it was because the method utilized the distribution differences of statistical histogram and statistical neighborhood histogram of the positive regions for noise removal, which did not consider the removal of the noise connected to the true positives. Therefore, it achieves a low p-rate, 31.01\%. Same as the results on CFD, \cite{rf29} could not give accurate crack locations, and achieves a low p-rate, 69.20\%. Suffering from the ``All Black'' issue, the FCN-VGG recognizes all crack and non-crack patches as background and produces all-black images as the results. For a fair evaluation, we conduct the comparison with the latest work DeepCrack on two different settings: one (DeepCrack-1) exactly follows the original paper trained with CrackTree data \cite{rf11}, and another (DeepCrack-2) re-trains the model using the industrial dataset. As present in Table II and Fig. 13, DeepCrack-1 introduces unacceptable noises due to the performance degradation on different domains as discussed at the beginning of this work; and DeepCrack-2 encounters the ``All Black'' problem. As discussed in section III-C, with the default settings, the discriminator in the original Pix2pix GAN will recognize both crack-GT and all-black-GT as real which damages the crack-GT generation ability and makes it like a regular U-Net; thus, it also produces all-black images as the results. By introducing the CPO-supervision and the adversarial learning with asymmetric U-Net generator, the model can be trained to generate crack-like results without losing the BG translation ability, and finally overcome the ``All Black'' issue. As shown in Fig. 13, it can detect thin cracks from the pavement images obtained from industrial settings. Refer Table II for quantitative results.

\begin{table}[t!]
	\centering
	\begin{center}
		\caption{Quantitative evaluation on dataset \cite{rf17}}
	\end{center}
	\label{tab:Eval-Data17}
	\begin{tabular}{|c|c|c|c|c|}
		\hline
		Methods & $P_{region}$ & $R_{region}$ & $F1_{region}$ & \textit{HD-score} \\ \hline
		CrackIT & \textbf{90.33\%} & 4.22\% & 8.06\% & 11 \\ \hline
		CrackForest & 36.21\% & \textbf{97.21\%} & 52.76\% & 65 \\ \hline 
		MPS\cite{rf17} & 79.01\% & 84.20\% & 81.52\% & 82 \\ \hline
		FCN-VGG \cite{rf7} & 0.00\% & 0.00\% & N/A & N/A \\ \hline
		Pix2pix GAN & 0.00\% & 0.00\% & N/A & N/A \\ \hline
		CrackGAN & 86.53\% & 94.20\% & \textbf{91.29}\% & \textbf{95} \\ \hline
	\end{tabular}
\end{table}

\begin{table}[htbp]
	\centering
	\caption{Comparisons of computational efficiency}
	\label{tab:Computation efficiency}
	\begin{tabular}{|c|c|c|c|}
		\hline
		Method & Time & Method & Time \\ \hline
		CrackIT & 6.1 s & DeepCrack & 2.4 s \\ \hline
		CrackForest & 4.0 s & Pix2pix GAN & 2.3 s \\ \hline
		\cite{rf29} & 10.2 s & \textbf{CrackGAN} & \textbf{1.6 s} \\ \hline
		FCN-VGG & 2.8 s &  &  \\ \hline
	\end{tabular}
\end{table}

\textbf{Dataset \cite{rf17}}: Fig. 14 and Table III present the results of CrackIT, CrackForest, MPS \cite{rf17}, FCN, Pix2pix GAN, and CrackGAN. Similar to the results on CGD, CrackIT missed quite a lot of cracks due to the drawback of feature extraction and post-processing. CrackForest could not remove the noises connected to the true crack regions and achieves a very low p-rate. MPS \cite{rf17} is a traditional image processing method based on minimal path selection; it performed the detection by following three steps, endpoint selection, minimal path estimation, and minimal path selection. It achieved good results as shown in Fig. 14 and Table III; however, it utilized quite a few tunable parameters and post-processing procedures that needed extra works manually. As discussed before, FCN-VGG and Pix2pix GAN fail the task due to the ``All Black'' issue. Instead, CrackGAN can properly handle the ``All Black'' problem and achieves the best performance.

\begin{figure}[t!]
	\centering
	\includegraphics[width=0.45\textwidth]{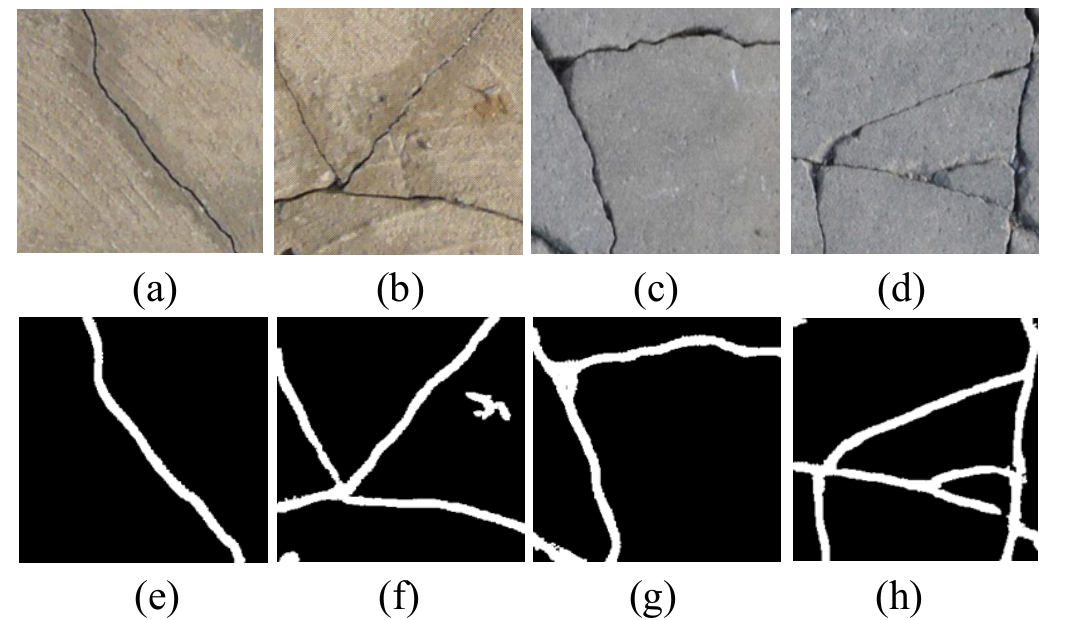}
	\caption{Crack detection on concrete pavement images and concrete wall images: (a) and (b) are concrete wall images; (c) and (d) are concrete pavement images; (e)-(h) are the detection results by CrackGAN.}
	\label{Fig15}
\end{figure}

In addition to pavement crack detection, the proposed method can also deal with other crack detection tasks; Fig. 15 provides crack detection results on concrete pavement images and concrete wall images based on the model trained with dataset \cite{rf7} and the labor-light GTs.

\subsection{Computational Efficiency}
In addition to the detection accuracy, it also compared the computation efficiency of the methods with public testing codes. The average processing times for processing a full-size image of 2048$\times$4096-pixel are present in Table IV. CrackIT-v1 \cite{rf58} takes 6.1 seconds based on a patch-wise processing; and CrackForest \cite{rf3} takes a relative less time (4.0 seconds) via using the parallel computing to implement the random forest for image patch classification. The two methods are implemented with Matlab-2016b on HP 620 workstation with 32G memory and twelve i7 cores. For the deep learning methods, they are implemented with the same computer but run on an Nvidia 1080Ti GPU with Pytorch. \cite{rf29} takes 10.2 seconds because it is based on the window-sliding. FCN \cite{rf7}, DeepCrack, Pix2pix GAN, and CrackGAN take much less time due to the FCN architecture; moreover, the CrackGAN takes much less time (i.e., 1.6 seconds) because it cuts off the last de-convolutional layer for the asymmetric U-Net design. 

%\IEEEpubidadjcol

\section{Conclusion}
In this work, we propose a novel deep generative adversarial network, named CrackGAN, for pavement crack detection. The method solves a practical and essential problem, ``All Black'' issue, existing in FCN-based pixel-level crack detection when using partially accurate GTs. More important, the network can solve crack detection tasks in a labor-light way. It can reduce the workload of preparing GTs significantly, and create the new idea for object detection/segmentation using partially accurate GTs. In addition, it can also solve the data imbalance problem which is the byproduct of the proposed approach. Moreover, the network is trained with small image patches, but can deal with any size images. The experiments demonstrate the effectiveness and superiority of the proposed method, and the proposed approach achieves state-of-the-art performance comparing with the recently published works.

Moreover, the theoretical analysis of neuron's property concerning receptive field can be employed to explain many phenomena in deep learning, such as the boundary vagueness in semantic segmentation \cite{rf6}, blurry of the generated images with GAN \cite{rf40, rf42}, etc., which have not been explained clearly yet. We believe that the analysis of each neuron’s property discussed in this paper could become a routine for designing effective neural networks in the future.

\ifCLASSOPTIONcaptionsoff
  \newpage
\fi



%\bibliographystyle{IEEEtran}


\begin{IEEEbiography}[{\includegraphics[width=1in,height=1.25in,clip,keepaspectratio]{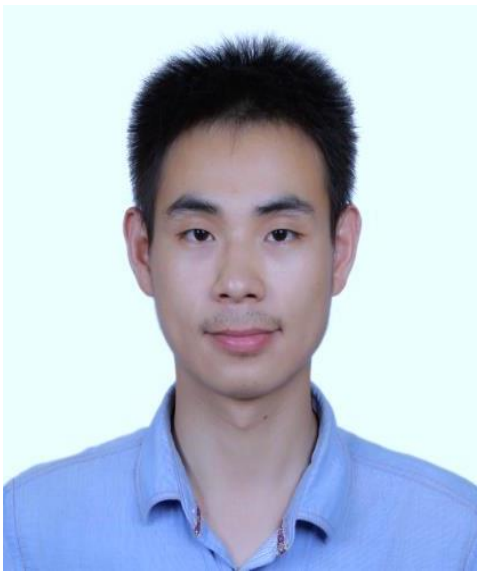}}]{Kaige Zhang}
received the B.S. degree in electronic engineering from Harbin Institute of Technology, Harbin, China, in 2011, and M.S. degree in signal and information processing from Harbin Engineering University, Harbin, China, in 2014. He is currently pursuing the Ph.D. degree in computer science with Utah State University. His research interests include computer vision, machine learning, and the applications on intelligent transportation systems, precision agriculture and biomedical data analytic.
\end{IEEEbiography}

\begin{IEEEbiography}[{\includegraphics[width=1in,height=1.25in,clip,keepaspectratio]{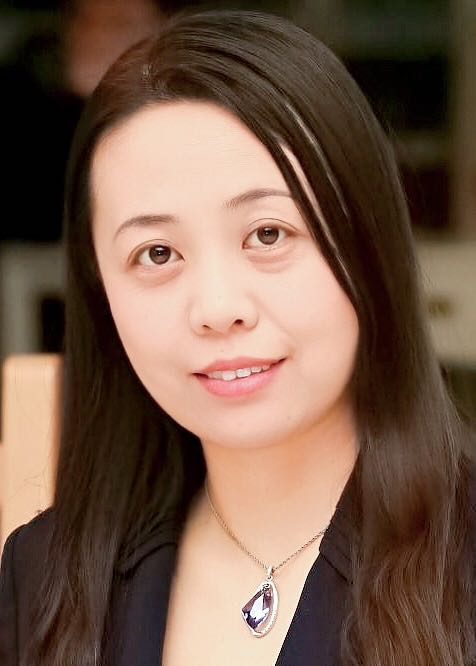}}]{Yingtao Zhang}
received her M.S. degree in Computer Science from Harbin Institute of Technology, Harbin, China, in 2004, and the Ph.D. degree in Pattern Recognition and Intelligence System from Harbin Institute of Technology, Harbin, China, in 2010. Now, she is an associate professor at School of Computer Science and Technology, Harbin Institute of Technology, Harbin, China. Her research interests include pattern recognition, computer vision, and medical image processing.
\end{IEEEbiography}

\begin{IEEEbiography}[{\includegraphics[width=1in,height=1.25in,clip,keepaspectratio]{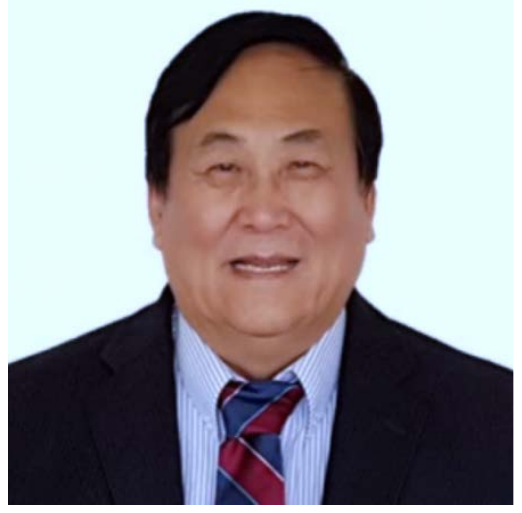}}]{Heng-Da Cheng}
received the Ph.D. degree in electrical engineering from Purdue University, West Lafayette, IN, in 1985, under the supervision Prof. K. S. Fu. He is currently a Full Professor with the Department of Computer Science, and an Adjunct Full Professor with the Department of Electrical Engineering, Utah State University, Logan, UT. He is an Adjunct Professor and a Doctorial Supervisor with the Harbin Institute of Technology. He is also a Guest Professor with the Institute of Remote Sensing Application, Chinese Academy of Sciences, Wuhan University, and Shantou University, and a Visiting Professor of Northern Jiaotong University, Huazhong Science and Technology University, and Huanan Normal University.

He has authored over 350 technical papers and is the Co-Editor of the book entitled \textit{Pattern Recognition: Algorithms, Architectures, and Applications} (World Scientific Publishing Company, 1991).

His research interests include image processing, pattern recognition, computer vision, artificial intelligence, medical information processing, fuzzy logic, genetic algorithms, neural networks, parallel processing, parallel algorithms, and VLSI architectures.

Dr. Cheng was the General Chair of the 11th Joint Conference on Information Sciences (JCIS) (2008), the tenth JCIS (2007), the Ninth JCIS (2006), and the Eighth JCIS (2005). He served as a Program Committee Member and the Session Chair for many conferences, and as a Reviewer for many scientific journals and conferences. He has been listed in Who's Who in the World, Who's Who in America, and Who's Who in Communications and Media.

Dr. Cheng is also an Associate Editor of \textit{Pattern Recognition}, \textit{Information Sciences}, and \textit{New Mathematics and Natural Computation}.

\end{IEEEbiography}


%\vfill


\end{document}